%% file: paper.tex
\documentclass[]{jingdong}
\usepackage[toc,page,header]{appendix}
\input{macros}
\input{sections/common}

\usepackage{cleveref}
\theoremstyle{plain}

\theoremstyle{definition}

\theoremstyle{remark}
\usepackage{subcaption}

\usepackage[textsize=tiny]{todonotes}
\usepackage{fontawesome}

\title{JoyAI-RA 0.1: A Foundation Model for\\Robotic Autonomy}
\author{Joy Future Academy, JD}

\affiliation{Full Author List in \hyperref[sec:contributions]{Contributions}}

\input{sections/abstract}

\checkdata[Project Page]{\url{https://joyai-ra.github.io/}}

\begin{document}
\maketitle

\input{sections/introduction}

\input{sections/relatedwork}

\input{sections/Approach}

\input{sections/experiments}

\newpage
\input{sections/conclusion}

\clearpage

\bibliographystyle{plainnat}
\bibliography{cite}

\clearpage
\input{sections/appendix}

\end{document}

%% file: macros.tex
\usepackage{minitoc}
\usepackage{url}
\usepackage{amssymb}
\usepackage[utf8]{inputenc}
\usepackage{microtype}
\usepackage{booktabs}
\usepackage{pifont} 
\usepackage{multirow}
\usepackage{makecell}
\usepackage{paralist}
\usepackage{xspace}
\usepackage{color}
\usepackage{xcolor}
\usepackage{colortbl}
\usepackage{adjustbox}
\usepackage{hyperref} 
\usepackage[edges]{forest}
\usepackage{tikz} 
\usepackage{caption}
\usepackage{amsfonts}
\usepackage[toc,page,header]{appendix}
\usepackage[utf8]{inputenc}
\usepackage{bbm}
\usepackage{enumitem}
\usepackage{pgfplots}
\pgfplotsset{compat=1.18}

\usepackage{subcaption}   % for subtable
\usepackage{pifont}

\usepackage{threeparttable}  % for g1 table
\usepackage{graphicx}  % same as above

\definecolor{seedc}{RGB}{7, 92, 173}

\newcommand{\name}[1]{GM-0}

\newcommand{\hardware}[1]{ByteMini}

\renewcommand{\paragraph}[1]{\vspace{0.1em}\noindent\textbf{#1}}

\usepackage{tabularx}       % 自动列宽/固定列宽

%% file: sections/common.tex
\usepackage{latexsym}
\usepackage[T1]{fontenc}
\usepackage[utf8]{inputenc}
\usepackage{microtype}
\usepackage{inconsolata}
\usepackage{graphicx}
\usepackage{hyperref}       % hyperlinks
\usepackage{url}            % simple URL typesetting
\usepackage{booktabs}       % professional-quality tables
\usepackage{amsfonts}       % blackboard math symbols
\usepackage{nicefrac}       % compact symbols for 1/2, etc.
\usepackage{stackengine}
\usepackage{microtype}      % microtypography
\usepackage{colortbl}
\usepackage{xcolor}
\usepackage{amsmath}
\usepackage{amssymb}
\usepackage{amsthm}
\usepackage{mathrsfs}
\usepackage{pifont}
\usepackage{MnSymbol}
\usepackage{balance}
\usepackage{enumitem}
\usepackage{listings}
\usepackage{xcolor}
\usepackage{natbib}
\usepackage{multicol}
\AtBeginDocument{%
  \providecommand\BibTeX{{%
    \normalfont B\kern-0.5em{\scshape i\kern-0.25em b}\kern-0.8em\TeX}}}

\makeatletter
\DeclareRobustCommand\onedot{\futurelet\@let@token\@onedot}
\def\@onedot{\ifx\@let@token.\else.\null\fi}

\usepackage{setspace}
\usepackage{mathtools}

\usepackage{multirow,booktabs}
\usepackage{subcaption}

\newcommand{\owo}[1]{\textsc{OAgents}}

\definecolor{lightgreen}{RGB}{144, 238, 144} 
\definecolor{lightred}{RGB}{255, 105, 97}

\newtcolorbox{promptbox}[2][Prompt]{
colback=black!5!white,
arc=5pt, 
boxrule=0.5pt,
fonttitle=\bfseries,
title=#1, 
before upper={\small}, fontupper=\fontfamily{ptm}\selectfont,
colframe=#2, % 使用传递的参数来设定 colframe
}
\definecolor{ogreen}{RGB}{34, 139, 34}

%% file: sections/abstract.tex
\abstract{

Robotic autonomy in open-world environments is fundamentally limited by insufficient data diversity and poor cross-embodiment generalization. Existing robotic datasets are often limited in scale and task coverage, while relatively large differences across robot embodiments impede effective behavior knowledge transfer. To address these challenges, we propose \textbf{JoyAI-RA}, a vision-language-action (VLA) embodied foundation model tailored for generalizable robotic manipulation. JoyAI-RA presents a multi-source multi-level pretraining framework that integrates web data, large-scale egocentric human manipulation videos, simulation-generated trajectories, and real-robot data. Through training on heterogeneous multi-source data with explicit action-space unification, JoyAI-RA effectively bridges embodiment gaps, particularly between human manipulation and robotic control, thereby enhancing cross-embodiment behavior learning. JoyAI-RA outperforms state-of-the-art methods in both simulation and real-world benchmarks, especially on diverse tasks with generalization demands.}

%% file: sections/introduction.tex
\section{Introduction}
\label{sec:introduction}

Achieving robust robotic autonomy in open-world environments requires manipulation policies that can be generalized to diverse tasks, scenes, and environments. Recent advances in large-scale robot learning and embodied foundation models have shown that scaling data and model capacity can substantially improve robot performance~\cite{brohan2022rt,driess2023palm,brohan2023rt,padalkar2023open}. In particular, vision-language-action (VLA) models offer a promising route to general-purpose robot control by coupling semantic understanding with action generation~\cite{black2410pi0,kim2024openvla,pi2025pi05,nvidia2025gr00t,team2025gemini}. However, open-world robotic autonomy remains bottle-necked by two tightly coupled challenges: insufficient data diversity for task coverage, and inevitable embodiment gap for multi-source behavior knowledge sharing.

Although recent efforts since RT-1 and Open X-Embodiment have significantly expanded the scope of robot learning~\cite{brohan2022rt,padalkar2023open}, collecting large amounts of high-quality robot interaction data is still expensive and operationally constrained. As a result, current corpora often underrepresent long-tail interactions, rare failure modes, and the diversity of scene layouts encountered in open-world settings. This lack of coverage restricts the breadth of behavior priors that policies can acquire and weakens robustness under distribution shift. Recent VLA systems have begun to explore broader forms of pretraining and cross-embodiment scaling~\cite{black2410pi0,pi2025pi05,being2025h05,galaxea2025g0,zheng2025uniact}, yet effective knowledge transfer across substantially different embodiments remains a central obstacle to scalable robot learning.

To step forward for addressing these challenges, we propose \textbf{JoyAI-RA}, a VLA embodied foundation model for generalizable robotic manipulation. JoyAI-RA leverages a large-scale multi-source pretraining dataset built from four complementary sources: multi-modal web data, egocentric human manipulation videos, simulation-generated trajectories, and real-robot demonstrations. Together, these data streams expand coverage across diverse tasks, scenes, and environments, providing the behavioral diversity necessary for open-world generalization. Built upon this heterogeneous corpus, JoyAI-RA designs a multi-level pretraining strategy with explicit action-space unification to alleviate embodiment gaps and enable effective knowledge transfer. Empirically, JoyAI-RA achieves strong performance on both simulation and real-world benchmarks, outperforming state-of-the-art approaches and demonstrating the value of multi-source heterogeneous pretraining for robotic autonomy.

In summary, this paper makes three main contributions. First, we propose JoyAI-RA, a VLA foundation model for generalizable robotic manipulation in open-world settings. Second, we design a multi-source, multi-level pretraining framework that unifies web data, egocentric human manipulation videos, simulation-generated trajectories, and real-robot data within a single training recipe. Third, we present that multi-source heterogeneous pretraining together with action-space unification effectively reduces embodiment gaps, improves cross-embodiment behavior transfer, and leads to strong downstream performance in both simulated and real-world evaluations.

%% file: sections/relatedwork.tex
\section{Related Work}
\paragraph{Multi-Source Data for Embodied Foundation Models.}
Recent progress in embodied foundation models has increasingly depended on combining heterogeneous data sources to provide sufficient semantic coverage, behavioral diversity, and embodiment-grounded supervision~\cite{khazatsky2024droid,walke2023bridgedata}.
Real-world robot demonstrations remain essential for grounding learning in executable behavior under sensing noise, contact uncertainty, and hardware constraints, as reflected in datasets such as AgiBot-World~\cite{bu2025agibot},
% Galaxea Open-World Dataset~\cite{jiang2025galaxea}, 
RoboMIND~\cite{hou2025robomind} and RoboCOIN~\cite{wu2025robocoin}.
RT-1~\cite{brohan2022rt} shows the value of scaling real-robot interaction data for policy learning, while Open X-Embodiment~\cite{padalkar2023open} extends this idea by aggregating robot demonstrations across many embodiments and collection setups.
Human videos have also become an important source of supervision, owing to their scalability of collection and alignment with real-world interaction patterns~\cite{hoque2025egodex,grauman2022ego4d,damen2018scaling,xperience_10m}. EgoDex~\cite{hoque2025egodex} further demonstrates that large-scale egocentric human videos can provide transferable priors for manipulation, especially when large-scale real-robot teleoperation data collection is costly.
Simulation data is commonly used to complement these data with scalable and controllable action supervision~\cite{tian2025interndata,li2023behavior,yin2026genie,contributors2025internroboticsrepo}. For example, InternData-A1~\cite{tian2025interndata} provides large-scale synthetic robot demonstrations across multiple embodiments for policy pretraining.
Building on these developments, JoyAI-RA propose our in-house human egocentric dataset EgoLive and our self-collected real-robot data JDAgibot.
% JoyAI-RA 
We explicitly structures multi-source data integration to exploit the complementary roles of real-robot, human, and simulation data for generalizable manipulation learning.

\paragraph{Universal Manipulation Policy Learning.}
% VLA 
Vision-language-action models have become a central paradigm for universal manipulation policy learning
% building generalist robot policies 
by coupling perception, language understanding, and action generation within a unified architecture. Representative systems include OpenVLA~\cite{kim2024openvla}, 
$\mathbf{\pi}_{0}$~\cite{black2410pi0}, 
$\mathbf{\pi}_{0.5}$~\cite{pi2025pi05}, GR00T N1~\cite{nvidia2025gr00t}, Gemini Robotics~\cite{team2025gemini}, etc. 
% and G0~\cite{galaxea2025g0}. 
In particular, LingBot-VLA~\cite{wu2026pragmatic} demonstrates that scaling a pragmatic VLA foundation model with large-scale real-world robot data can yield strong generalization performance across diverse tasks and robotic platforms. Beyond simply scaling model size and training data, recent works have also begun to more systematically explore the broader design space of VLA-based policy learning.
RoboVLMs~\cite{li2024robovlms} studies how backbone choices, policy formulation, and cross-embodiment data mixtures affect generalist policy learning, while VLA-Adapter~\cite{wang2024vladapter} and SpatialVLA~\cite{qu2025spatialvla} further highlight the importance of efficient adaptation and spatial representations for action prediction. Together, these works underscore the growing need for architectures that can balance generalization, efficiency, and structured perception in embodied settings. At the policy level, Motus~\cite{bi2025motus} formulates a Mixture-of-Transformer architecture that jointly supports understanding, video generation, and action prediction, while ABot-M0~\cite{yang2026abot} further improves policy generation through cross-embodiment standardization, geometry-aware perception, and action manifold learning. Compared with flow-based policies such as $\mathbf{\pi}_{0}$ and $\mathbf{\pi}_{0.5}$, JoyAI-RA further emphasizes a structured multi-stage training recipe that combines broad multi-modal co-pretraining, cross-embodiment action alignment, and a dedicated Perceiver-based action expert for continuous action generation, rather than relying primarily on scaling continuous robot control alone.

%% file: JoyAI-RA/sections/approach.tex
\begin{figure}[t]
    \centering
    \includegraphics[width=0.9\linewidth]{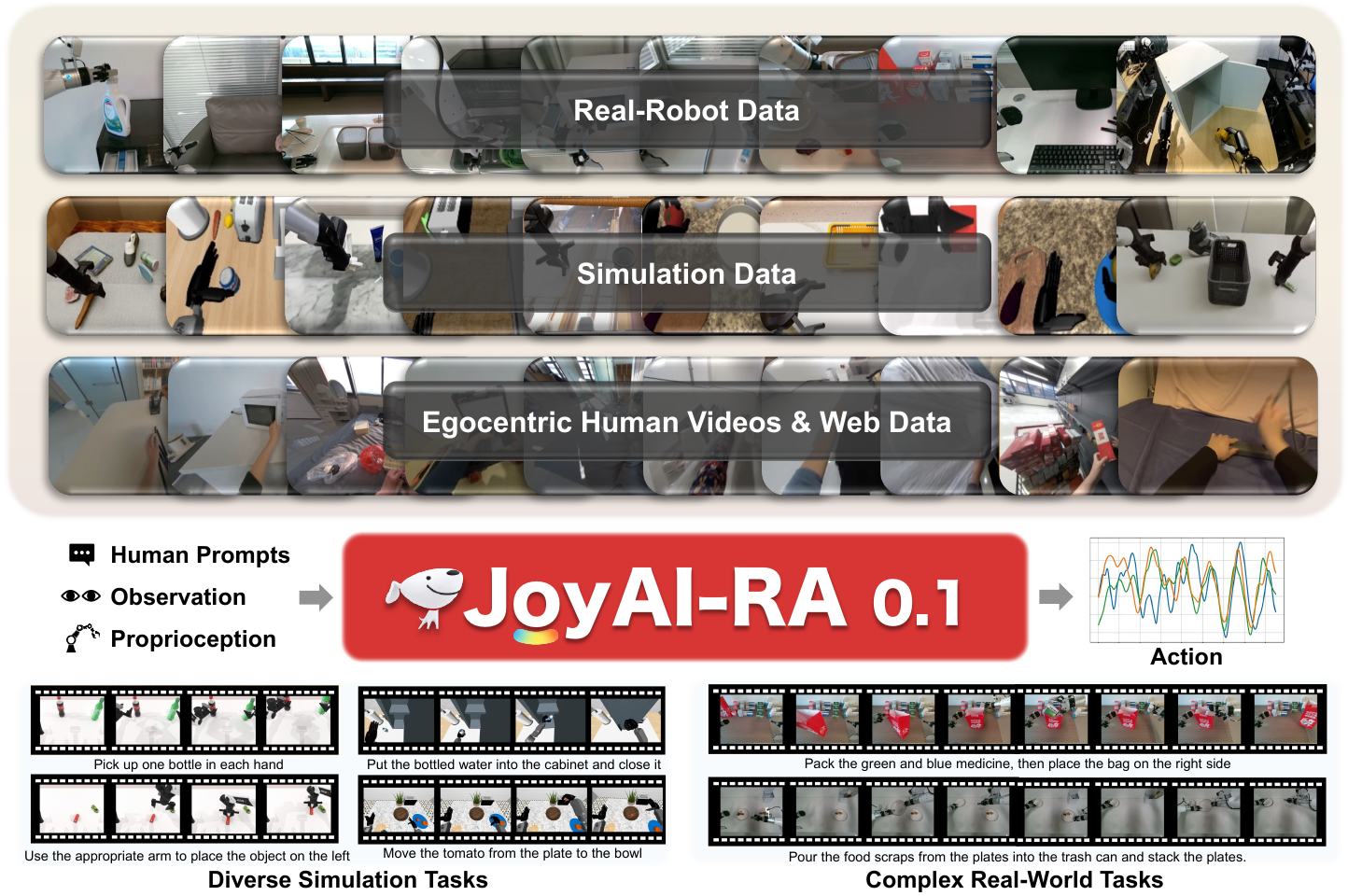}
    \caption{ Overview. JoyAI-RA is trained on four complementary data sources: web data, egocentric human manipulation videos, simulation-generated trajectories, and real-robot demonstrations. Through a multi-level pretraining framework with action-space unification, JoyAI-RA learns a unified VLA policy that generalizes across diverse and complex tasks.}
    \label{fig:overview}
    \vspace{-5pt}
\end{figure}

\section{Approach}
\label{sec:approach}

\subsection{Overview}

Figure~\ref{fig:overview} presents an overview of JoyAI-RA 0.1, which integrates diverse data sources, including real-robot demonstrations, large-scale simulation data, and egocentric human videos and web data, within a unified training pipeline. Given multimodal inputs consisting of language instructions, visual observations, and proprioceptive states, JoyAI-RA predicts temporally consistent action sequences for downstream control across diverse simulation tasks and complex real-world tasks.
We describe these data sources in Sec.~\ref{sec:dataset}. They provide complementary supervision in both scale and structure, allowing the model to learn from rich semantic diversity, structured interaction environments, and embodiment-aware execution.

To bridge the gap across data sources and embodiments, we introduce a unified action space (Sec.~\ref{sec:unified action space}) that projects both proprioceptive states and actions into a shared action representation with fixed action dimensionality. This ensures consistent physical semantics across robot platforms, and flexibly accommodates heterogeneous robot morphologies via action masking, collectively facilitating cross-embodiment knowledge transfer across heterogeneous data sources.
To effectively learn from heterogeneous data, we design the model architecture in Sec.~\ref{sec:model} and adopt a multi-stage training paradigm in Sec.~\ref{sec:training recipe}, enabling stable cross-modal learning and improving both generalization and execution performance.

Together, these components form a cohesive framework that equips JoyAI-RA to scale across heterogeneous data sources and generalize to diverse tasks and robot embodiments.

\subsection{Dataset}
\label{sec:dataset}

\begin{figure}[t]
    \centering
    \includegraphics[width=0.9\linewidth]{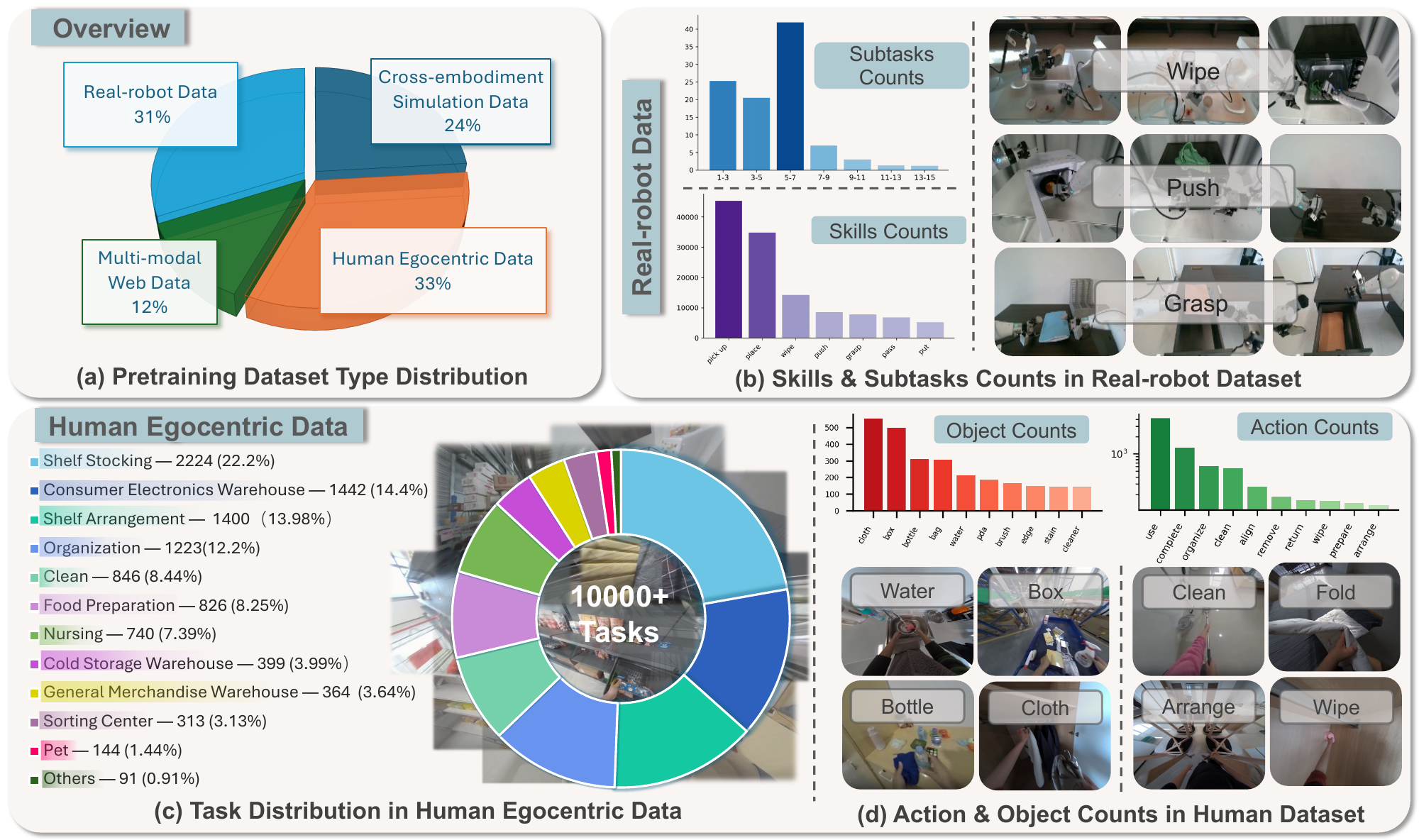}
    \caption{The distribution analysis of pretraining data. (a) Composition of pretraining data, including cross-embodiment, simulation, human, and web sources. (b) Robot dataset statistics, showing distributions of subtasks and manipulation skills. (c) Human dataset task distribution across diverse domains, totaling over 10k tasks. (d) Human dataset statistics of action types and object categories.}
    \label{fig:Distribution analysis}
    \vspace{-5pt}
\end{figure}

We organize our training datasets of JoyAI-RA from four data sources, as shown in Figure~\ref{fig:Distribution analysis}(a).
% , including Multi-modal Web Data, Human Egocentric Data, 
These data sources provide complementary supervision, ranging from semantic and perceptual priors to embodiment-aware action learning, and enable our JoyAI-RA to bridge language understanding and real-world robot execution.

\textbf{Multi-Modal Web Data.}
% \textbf{Visual Language Data.}
We incorporate large-scale visual instruction-tuning data as the semantic layer of our embodied multi-source data to boost JoyAI-RA with strong perceptual and linguistic priors before action learning.
Specifically, we collect a multi-modal dataset containing visual question-answering pairs from several general-purpose and spatially grounded resources, such as Cambrian-10M~\cite{tong2024cambrian}, RefSpatial~\cite{zhou2025roborefer}, Galaxea~\cite{jiang2025galaxea}, and Cosmos-Reason1-SFT~\cite{azzolini2025cosmos}.
Although these datasets do not provide executable manipulation trajectories, they substantially enrich the model's understanding of task-relevant semantics and language-conditioned perception, both of which are essential for open-world manipulation.
% robotic manipulation.
% manipulation tasks

%TODO 女娲数据命名
\textbf{Egocentric Human Manipulation Data.} We construct an in-house human dataset, \textbf{EgoLive}, consisting of egocentric videos captured with RGB cameras at 60 FPS, as illustrated in Figure~\ref{fig:Distribution analysis} (c) and (d).
The dataset provides broad and diverse coverage, spanning 1,969 object categories and 1,796 action categories, significantly expanding beyond prior datasets such as EgoDex~\cite{hoque2025egodex}
% and introducing over one hundred 
with more than 100 unique categories. It covers several tasks from a wide range of real-world scenarios, including 
% household, retail, logistics, and automotive maintenance domains, with 
3,779 tasks in household settings, 3,686 tasks in retail environments, and 2,518 tasks in logistics scenarios. Moreover, the dataset encompasses both long-horizon and short-horizon tasks, enabling the study of hierarchical planning and multi-step manipulation behaviors.

Compared to existing datasets, our dataset places a stronger emphasis on operation-centric textual annotations. In addition to 
% providing 
high-level task descriptions for each episode, it
% further 
includes fine-grained per-frame subtask annotations, offering rich supervision for learning temporally grounded action decomposition and sequential reasoning. To process these data, we develop in-house hand-pose estimation pipelines to recover human hand trajectories, which are subsequently retargeted to multiple robot embodiments, including
% platforms such as 
ALOHA~\cite{zhao2023learning}, Fourier~\cite{fourier_intelligence} and Agibot G1 robots. While large-scale data collection inevitably introduces noise, the combination of high frame rate, extensive diversity, fine-grained annotations, and cross-embodiment action alignment makes our dataset
% to serve as
a strong foundation for learning generalizable and transferable manipulation policies.

\textbf{Simulation Data.}
Simulation data serves as a critical foundation for robotic action learning, bridging broad semantic pretraining and embodiment-grounded real-world policy learning.
We leverage 
several synthetic robotic datasets for action learning, including InternData-A1~\cite{tian2025interndata}, GenieSim3.0-Dataset~\cite{yin2026genie}, and InternData-M1~\cite{contributors2025internroboticsrepo}, and select curated subsets from them for training.
As a scalable source of
% Simulation data serves as a scalable source of 
embodiment-aware supervision, simulation data facilitates the transfer from human-centric and vision-language pretraining to robot action learning.

% while also helping reduce embodiment gaps across heterogeneous datasets.

\begin{comment}
Simulation data for robotics is a key foundation for action learning, bridging broad semantic pretraining and embodiment-grounded real-world policy learning.
% which provides strong supervision for robot action learning and offers guidance for learning from human data.
We propose a heterogeneous collection of \textbf{???} hours of high-quality synthetic robot demonstrations, including 
% AgiBot-World~\cite{bu2025agibot},
InternData-A1~\cite{tian2025interndata}, etc.,
% This collection 
and covering \textbf{???} robot embodiments, such as Split Aloha, Fourier GR1, and Agibot G1.
% , reflecting diverse robot structures and task settings. 
Simulation data serves as a scalable source of embodiment-aware supervision that facilitates transfer from human-centric and vision-language pretraining, and helps reduce embodiment gaps with heterogeneous datasets.

\end{comment}

\textbf{Real-Robot Data.}
Our real-robot data collection integrates open-source multi-embodiment datasets—including the Open-X-Embodiment~\cite{padalkar2023open}, AgiBot-World~\cite{bu2025agibot}, and Galaxea Open-World Dataset~\cite{jiang2025galaxea}—alongside our in-house self-collected dataset, denoted \textbf{JDAgibot}.
Compared with simulated robotic demonstrations, real-robot trajectories capture contact uncertainty, sensing noise, and hardware constraints that arise in physical deployment, thereby improving the learning of robust policies that transfer to the real world.

\subsection{Unified Action Space}
\label{sec:unified action space}

Effectively leveraging heterogeneous multi-source data for VLA policy training requires reconciling the incompatible action representations that arise from diverse embodiments and data sources.
To address this, we define a unified action space that converts both proprioceptive states and actions into a common physical semantics space with a consistent dimensionality, enabling the model to learn transferable knowledge across heterogeneous data sources.

\textbf{Camera-Frame End-Effector Representation.}
Across action-labeled data, including real-robot demonstrations, retargeted human trajectories, and simulation, we represent end-effector states and the corresponding actions in the camera coordinate frame whenever reliable extrinsics and end-effector pose supervision support a well-defined camera-relative parameterization.
Concretely, the 6-DoF end-effector pose is decomposed into a 3-dimensional translation vector and a 3-dimensional axis-angle rotation vector. 
Expressing actions in the camera frame, rather than in a robot-specific base frame or joint-level configuration, provides two key benefits.
First, it confers consistent physical semantics: a given action vector encodes the same spatial displacement and rotation regardless of the robot's base placement or kinematic configuration.
Second, it naturally aligns with the visual observations fed to the VLM backbone, since both the image inputs and the predicted actions share the same viewpoint, facilitating visually grounded action prediction.

\textbf{Unified Action Dimensionality.}
Beyond coordinate alignment, cross-embodiment training requires a fixed-length action representation that can accommodate robots with substantially different kinematic structures.
We define a unified action vector that covers all actuator groups appearing across our data sources, including the left and right arms, the left and right dexterous hands, and the left and right grippers, etc. For each embodiment, unified-vector dimensions without a corresponding realized degree of freedom are masked in the loss and omitted from gradients, so the same fixed-dimensional representation can train across morphologies from single-arm grippers to bimanual dexterous-hand systems.

\subsection{Model Architecture}
\label{sec:model}

\begin{figure}[t]
    \centering
    \includegraphics[width=0.95\linewidth]{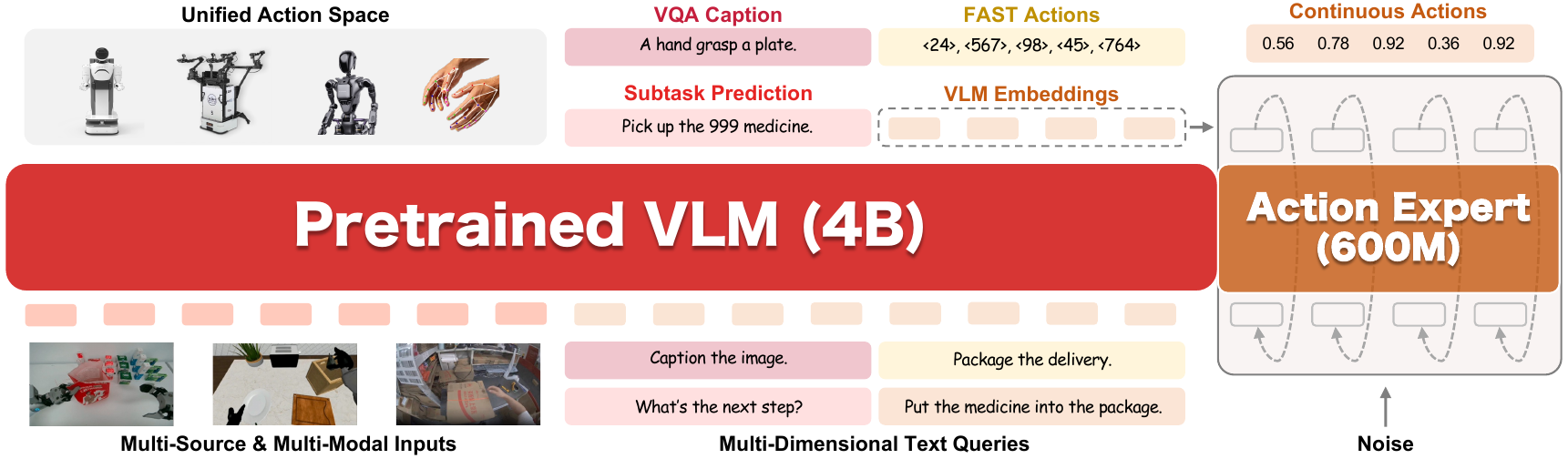}
    \caption{The model architecture of JoyAI-RA.}
    \label{fig:Model Architecture}
    \vspace{-10pt}
\end{figure}

Figure~\ref{fig:Model Architecture} illustrates JoyAI-RA's modular design: a vision-language model (VLM) handles vision-language understanding, and a perception-action expert, built upon a Perceiver~\cite{jaegle2021perceiver} architecture for efficient multi-modal fusion via latent bottlenecks, generates actions, decoupling semantics from low-level control. Given multi-view observations and language inputs, the VLM produces spatially grounded multimodal representations that encode both semantic and geometric context. These representations are provided as contextual inputs to the perception-action expert, which predicts temporally consistent continuous actions.

JoyAI-RA jointly models tokenized textual outputs and continuous action chunks. Given multi-view observations $O_t$ and a language instruction $L$, the model predicts both a textual output $\ell$ (\textit{e.g.}, a subtask) and a sequence of low-level actions $a_{t:t+H}$. In practice, the model first generates a high-level semantic description and then produces corresponding action chunks conditioned on the observation and this intermediate representation.

To enable stable continuous control, we formulate the perception-action expert as a conditional velocity field predictor under a flow-matching framework, where the model learns to estimate the velocity of noisy action latents. The latent sequence is constructed by concatenating proprioceptive state, learnable future tokens (\textit{i.e.}, a set of trainable query embeddings that represent placeholders for future action steps), and timestep-conditioned noisy action features along the token dimension:
\[
a^{0}_{t:t+H} = \mathrm{Concat}\!\left(\phi_s(s_t),\, f_{\mathrm{future}},\, \phi_a(\tilde{a}_{t:t+H}, \tau)\right),
\]
where $\phi_s(\cdot)$ and $\phi_a(\cdot)$ denote the state and action encoders, respectively, and $\tilde{a}_{t:t+H}$ represents the noisy action trajectory obtained by interpolating between Gaussian noise and the ground-truth action chunk $a_{t:t+H}$ at timestep $\tau$ during training. Given the visual-language representation $z_t$, action features $ a^{0}_{t:t+H}$, and timestep $\tau$, the expert predicts a velocity field over the action chunk:
\[
v^{\text{out}}_{t:t+H} = f_{\theta}(z_t, a^{0}_{t:t+H}, \tau).
\]
We instantiate $f_{\theta}$ as a stack of time-aware Perceiver attention blocks. At each layer, timestep-adaptive normalization is first applied to the visual-language stream, $\tilde{z}_t = \mathrm{AdaLN}_z(z_t, \tau)$, enabling explicit modulation of perceptual features by the denoising timestep. The latent tokens are then updated via a residual attention mechanism:
\[
h'_{t:t+H} = h_{t:t+H} + \mathrm{MHA}\!\left(
Q = h_{t:t+H}, \;
K,V = [h_{t:t+H}; \tilde{z}_t]
\right),
\]
followed by a residual feed-forward refinement $h^{\text{out}}_{t:t+H} = h'_{t:t+H} + \mathrm{MLP}(h'_{t:t+H})$. The final velocity prediction $v^{\text{out}}_{t:t+H}$ is decoded from the refined latent representation.This explicit temporal conditioning allows the model to adapt its velocity predictions across different denoising stages, improving temporal consistency and yielding more stable action generation for embodied manipulation.

\subsection{Training Recipe}
\label{sec:training recipe}

To more effectively learn and consolidate knowledge across heterogeneous data sources, we structure training as three phases that progressively specialize while preserving cross-source transfer.
The first, VLM Co-Pretraining, builds broad vision-language and embodied priors from diverse multimodal supervision.
The second, VLA Co-Pretraining, centers on learning continuous manipulation behaviors through action prediction, while preserving the backbone's visual-semantic representations.
Finally, VLA Post-Training lightly fine-tunes on a small domain-specific corpus to boost downstream performance while retaining multi-source transfer.

\subsubsection{VLM Co-Pretraining}

During the VLM Co-Pretraining stage, we train the VLM model on a diverse mixture of data sources with task-specific sampling ratios, aiming to endow it with visual understanding, spatial reasoning, long-horizon planning, and tokenized action prediction capabilities. The training data consists of four main categories: 
(1) General VQA: The original visual comprehension ability is preserved through common formats such as QA pairs, visual captioning, and multi-turn dialogue.
(2) Embodied VQA: By incorporating multimodal data that includes embodied capability terms—comprising both image and video inputs, and covering output formats such as points, bounding boxes, and trajectories—the model’s spatial understanding and reasoning abilities are enhanced. Additionally, a portion of the task decomposition data is included to strengthen long-horizon planning capabilities.
(3) Cross-embodiment action data: Using both real and simulated action data, the model is initially equipped with the ability to generate trajectories through discrete action generation, which serves as a foundation for subsequent continuous action generation.
(4) Human video data: We incorporate human video data to obtain a wider range of visual inputs and richer action distributions, thereby improving the model’s generalization ability across different scenarios.

% \textbf{Training objective}
In the VLM co-pretraining stage, the standard training objective for autoregressive architectures is to minimize the negative log-likelihood of the target tokens: 
\begin{equation}
\mathcal{L}_{\text{VLM Co-Pretraining}}(\theta) = \mathbb{E}_{(x, y) \sim \mathcal{D}} \left[ -\log p_{\theta}(y_{1:n} \mid x_{1:n}) \right] = \mathbb{E}_{(x, y) \sim \mathcal{D}} \left[ -\sum_{j=1}^{n-1} M_j \log p_{\theta}(y_{j+1} \mid x_{1:j}) \right]  
\end{equation}

where $x$ and $y$ denote input tokens and output tokens, respectively. Notably, $y$ encompasses both VLM responses and discrete FAST~\cite{pertsch2025fast} action tokens. $M$ represents a loss mask specifying the tokens to be predicted, and $\mathcal{D}$ is the VLM pretraining dataset.

\subsubsection{VLA Co-Pretraining}

% \textbf{Training Data} 
In the VLA co-pretraining phase, two primary data types, General VQA and Embodied VQA, are progressively introduced to maintain the model's basic competencies and embodied spatial reasoning abilities. For action supervision, we use simulated robot trajectories, real-robot demonstrations, and human video demonstrations retargeted to the downstream embodiment, all represented in the unified action space so heterogeneous motion sources remain aligned.

% \textbf{Training objective}
In addition to the autoregressive loss used in the first stage, the introduction of the action expert necessitates the computation of an additional flow matching loss. The total loss is thus given by:
\begin{equation}
\mathcal{L}_{\text{VLA Co-Pretraining}}(\theta) = \alpha * \mathbb{E}_{(x, y) \sim \mathcal{D}} \left[ -\sum_{j=1}^{n-1} M_j \log p_{\theta}(y_{j+1} \mid x_{1:j}) \right]  +  \mathbb{E}_{\mathcal{D}, \tau, \omega} \left[ \| \omega - a_{1:H} - f_\theta^a(a^{\tau, \omega}_{1:H}) \|^2 \right],
\end{equation}

given the flow matching time index $\tau \in [0, 1]$, the input to the model takes the form of a perturbed action chunk 
$a^{\tau, \omega}_{1:H} = \tau a_{1:H} + (1 - \tau) \omega$, where $\omega$ is sampled from $\mathcal{N}(0, I)$. 
The training objective is to predict the flow $\omega - a_{1:H}$, $f_\theta^a$ represents action flow prediction expert.  $\alpha$ is a loss multiplier that balances the action prediction via flow-matching against the standard vision-language modeling loss. 

\subsubsection{Post-Training on Target Robots}

% \textbf{Training Data}
In the post-training stage, only data collected from the target robot embodiment is used, so that the pretrained policy can be further specialized to the downstream deployment setting. At this stage, we discard the auxiliary autoregressive objective used in the co-pretraining stages and optimize the model solely with the flow matching objective over continuous target-robot actions. The objective is defined as:
\begin{equation}
\mathcal{L}_{\text{Post-Training}}(\theta) = \mathbb{E}_{\mathcal{D}_{\text{target}}, \tau, \omega} \left[ \| \omega - a_{1:H} - f_\theta^a(a^{\tau, \omega}_{1:H}) \|^2 \right],
\end{equation}

where $\mathcal{D}_{\text{target}}$ denotes the post-training dataset collected on the target robot or generated from the target simulation benchmark, $\tau \in [0, 1]$ is the flow matching time index, and $\omega \sim \mathcal{N}(0, I)$ is the sampled Gaussian noise. During post-training, all model parameters are updated end-to-end, allowing the visuomotor representations and action generation module to adapt jointly to the target tasks and thus improve the success rate in both real-world deployment and the corresponding simulation evaluations.

%% file: sections/experiments.tex
\section{Experiments}
\label{sec:experiments}

\subsection{Experimental Setup}
% \textbf{Simulation Benchmarks.} 
We evaluate JoyAI-RA on simulation and real-world benchmarks to assess its effectiveness and generalization across diverse robot embodiments.

\textbf{Simulation Benchmark.} We evaluate JoyAI-RA on simulation benchmarks including RoboTwin 2.0~\cite{chen2025robotwin} and Robocasa GR1 Tabletop tasks~\cite{nvidia2025gr00t}. For RoboTwin 2.0, we conduct a multi-task training setup following Motus~\cite{bi2025motus} where all models are trained on 2,500 demonstrations collected in clean scenes (50 per task) plus 25,000 demonstrations from heavily randomized scenes (500 per task). The randomizations include random backgrounds, table clutter, table height perturbations, and random lighting. We evaluate under both Easy (fixed initial configurations) and Hard (varied object poses and scene layouts) settings. For the Robocasa GR1 Tabletop tasks, we train a single model on all 24 tasks following starVLA~\cite{community2026starvlalegolikecodebasevisionlanguageaction}, with mean results averaged over 50 rollouts per task. This is a 6DoF dexterous hand benchmark that extends RoboCasa~\cite{nasiriany2024robocasa} with task-specific environments and assets. It comprises 24 manipulation tasks involving articulated objects, designed to evaluate general-purpose robot policies in household scenarios.

\begin{figure}[t] % [htbp] 表示建议放置在：当前位置、页面顶部、页面底部或单独一页
    \centering % 使图片居中
    \includegraphics[width=0.95\linewidth]{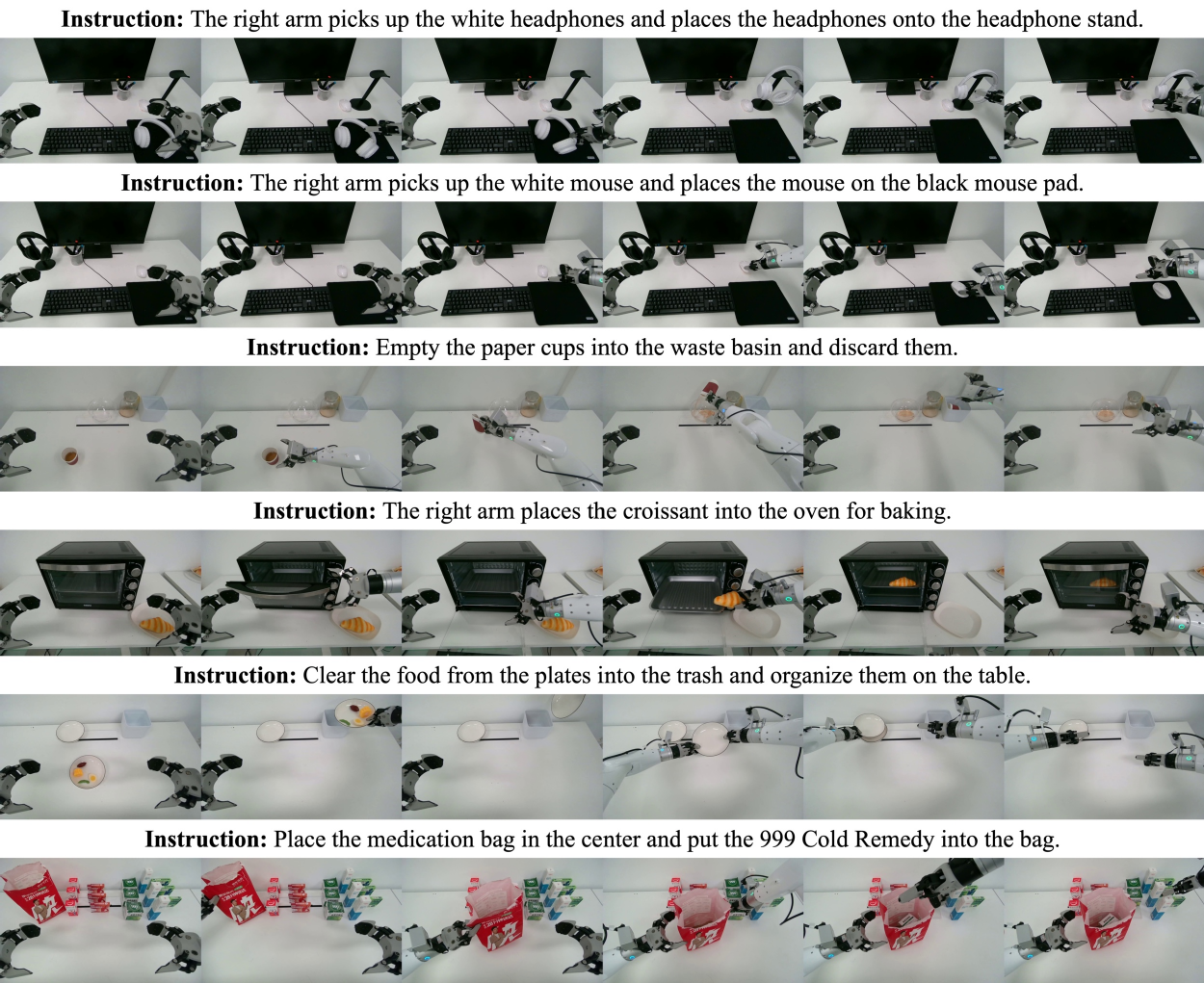} % 设置宽度为正文宽度，并指定文件名
    \caption{An overview of real-world Agibot benchmark from 5 scenarios on the AgiBot G1 platform.} % 这里写图片的标题
    \label{fig:001-006} % 这里写引用标签，方便在正文中用 \ref{fig:001-006} 引用
    \vspace{-10pt}
\end{figure}

\textbf{Real-World Agibot Benchmark.} To systematically evaluate JoyAI-RA on real-world robotic manipulation, we conduct Real-World Agibot Benchmark on the AgiBot G1 humanoid robot platform. We design 5 representative real-world scenarios, including office, tea table, kitchen, dining table, and pharmacy, covering a total of 6 evaluation tasks in Figure \ref{fig:001-006}. For each evaluation task, we evaluate the model over 20 trials. In order to assess the generalization of the model, we vary the instruction, as well as the position, orientation, color, and texture of the manipulated objects and the environmental configurations within reasonable ranges. Detailed task descriptions are provided in Appendix Section~\ref{sec:appendix_real_world}.

\input{table/sub_robotwin_full}

\input{table/sub_robocasa}

% \newpage

\subsection{Results}
\subsubsection{RoboTwin 2.0}

Table~\ref{tab:sub-robotwin-short} summarizes the results on RoboTwin 2.0. JoyAI-RA consistently outperforms all strong baselines under both the Easy and Hard settings, achieving average success rates of \textbf{90.48\%} and \textbf{89.28\%}, respectively, while reaching a 100\% success rate on several representative tasks such as \textit{Adjust Bottle}, \textit{Grab Roller}, and \textit{Place Empty Cup}. These results suggest that the proposed unified action space and multi-stage training paradigm enable effective modeling of diverse manipulation patterns, including fine-grained grasping, symmetric bi-manual coordination, and dynamic trajectory control. Moreover, the consistent improvements over prior methods, particularly under diverse initial conditions, indicate that leveraging heterogeneous multi-source data leads to transferable manipulation priors that substantially enhance both generalization and robustness under heavy scene randomization.

\subsubsection{RoboCasa GR1 Tabletop}

The evaluation results on RoboCasa GR1 Tabletop~\cite{nasiriany2024robocasa} tasks are presented in Table~\ref{tab:sub-robocasa-results}. JoyAI-RA achieves a new state-of-the-art average success rate of \textbf{63.2\%}, outperforming all prior methods by a clear margin, with particularly notable gains on complex, long-horizon tasks such as \textit{CanToDrawerClose} (+16.0 over the previous best), \textit{MilkToMicrowaveClose} (+24.0), and \textit{TrayToPot} (+18.0). These improvements highlight the effectiveness of JoyAI-RA in handling sequential reasoning and precise object manipulation, and further demonstrate that the manipulation priors learned from heterogeneous multi-source data generalize well across diverse task distributions, resulting in robust performance in challenging, compositional environments.

\subsubsection{Real-World AgiBot Benchmark}

\begin{figure}[t] % [htbp] 表示建议放置在：当前位置、页面顶部、页面底部或单独一页
    \centering % 使图片居中
    \includegraphics[width=0.9\linewidth]{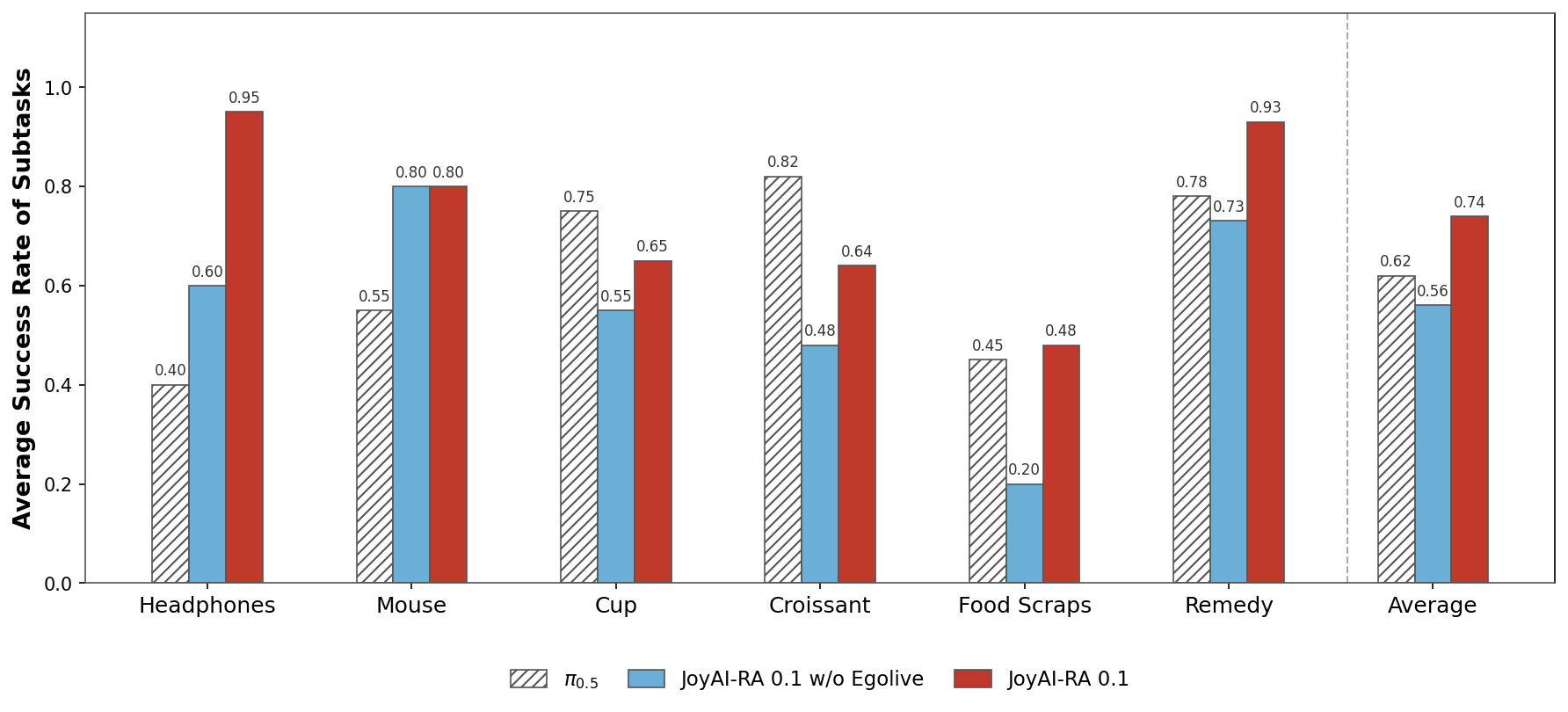} % 设置宽度为正文宽度，并指定文件名
    \vspace{-10pt}
    \caption{Main evaluation results on real-world Agibot benchmark.} % 这里写图片的标题
    \label{fig:real_robot_evaluation_results} % 这里写引用标签，方便在正文中用 \ref{fig:real_robot_evaluation_results} 引用
    %\vspace{-10pt}
\end{figure}

The main results on the real-world AgiBot benchmark are shown in Figure~\ref{fig:real_robot_evaluation_results}, which reports the average success rate across six household tasks. Compared with $\pi_{0.5}$~\cite{pi2025pi05}, JoyAI-RA achieves stronger overall real-robot performance, improving the cross-task average from 0.62 to \textbf{0.74}. Observed under diverse real-world deployment conditions, this gain supports the effectiveness of our multi-source, multi-level pretraining framework in enabling cross-embodiment knowledge transfer.

At the task level, JoyAI-RA shows the largest margins on \textit{Headphones} and \textit{Remedy}, where accurate target recognition, instruction following, and precise final placement are critical, and it also improves over $\pi_{0.5}$ on \textit{Mouse} and \textit{Food Scraps}. In contrast, $\pi_{0.5}$ remains higher on \textit{Cup} and \textit{Croissant}, suggesting that long-horizon, precise visual reasoning, and graded sequential manipulation are still challenging for our model despite gains elsewhere; \textit{Food Scraps} in particular stays difficult for all methods in absolute terms. Overall, JoyAI-RA delivers stronger real-world manipulation capability, with especially clear benefits on semantically grounded placement while structured multi-step tasks indicate directions for future improvement.

\input{table/sub_tasks_table}

\begin{figure}[t] % [htbp] 表示建议放置在：当前位置、页面顶部、页面底部或单独一页
    \centering % 使图片居中
    \includegraphics[width=0.9\linewidth]{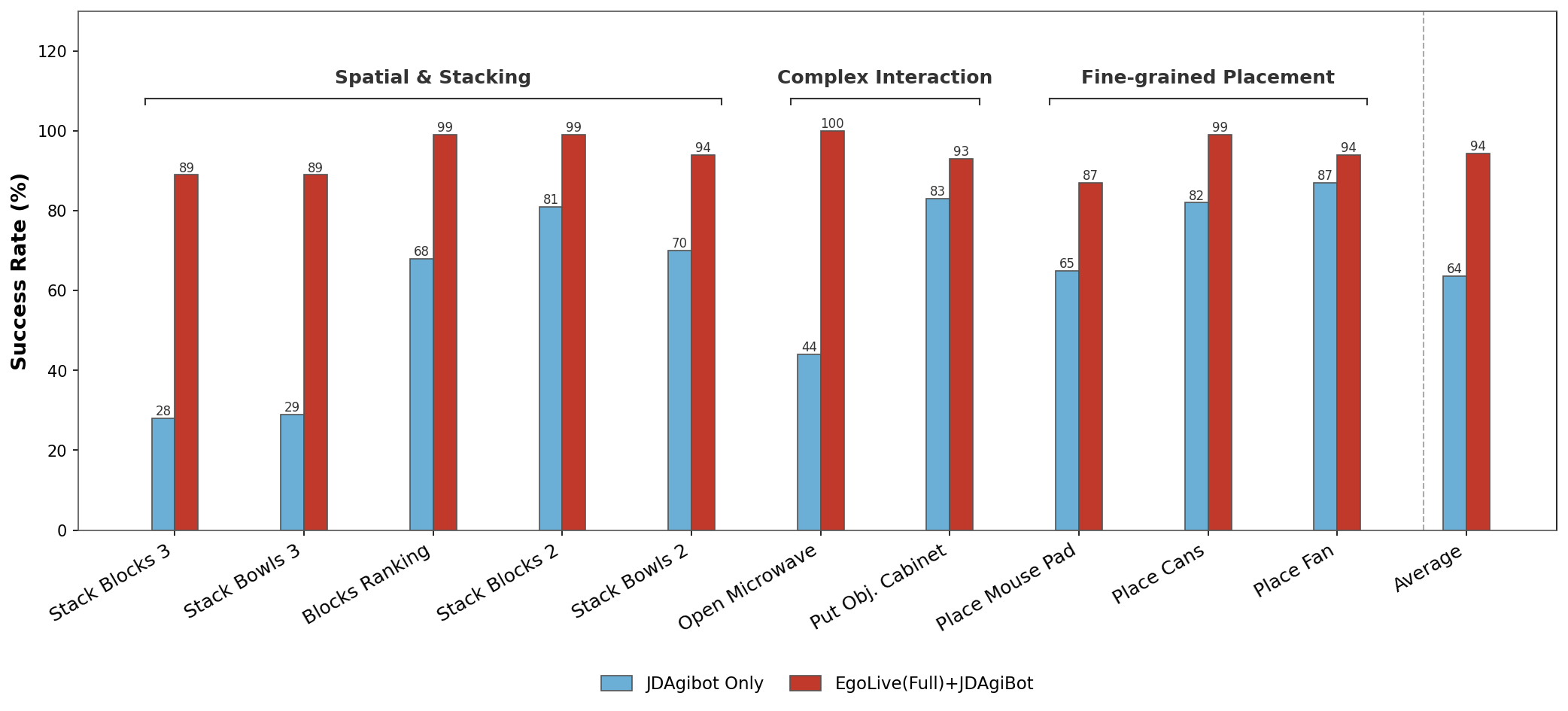} % 设置宽度为正文宽度，并指定文件名
    \vspace{-10pt}
    \caption{Performance comparison of representative tasks between JDAgiBot Only and EgoLive-1000$+$JDAgiBot on the RoboTwin 2.0 benchmark. Tasks are grouped into three categories: Spatial \& Stacking, Complex Interaction, and Fine-grained Placement.} % 这里写图片的标题
    \label{fig:robotwin50_representative} % 这里写引用标签，方便在正文中用 \ref{fig:robotwin50_representative} 引用
\end{figure}

%\subsection{Ablation Study on EgoLive Data}
\subsection{Effectiveness Analysis of EgoLive Data}

To further verify the effectiveness of our proposed egocentric human video data (i.e., \textit{EgoLive}), we conduct ablations on the RoboTwin 2.0 simulation and real-world AgiBot benchmark. For RoboTwin 2.0, all ablations use VLA pretraining only: we do not apply a separate VLM-only pretraining stage, and we compare four regimes for VLA pretrain data---\textit{No Pretraining}, \textit{JDAgibot Only}, \textit{JDAgibot + 10\% EgoLive}, and \textit{JDAgibot + Full EgoLive}. For the real-world AgiBot benchmark, we compare two pretrained checkpoints that differ only in whether EgoLive is included.

\begin{figure}[t] % [htbp] 表示建议放置在：当前位置、页面顶部、页面底部或单独一页
    \centering % 使图片居中
    \includegraphics[width=0.9\textwidth]{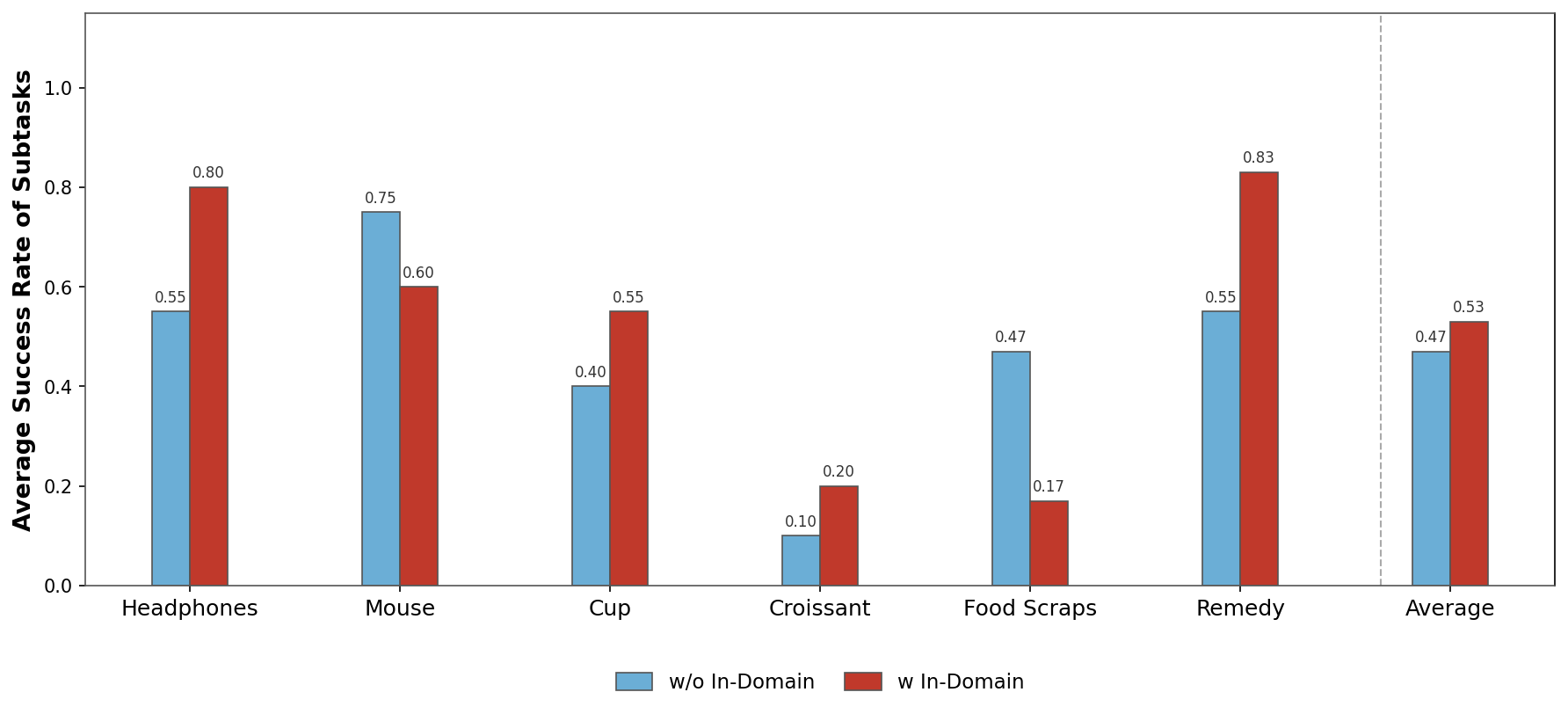} % 设置宽度为正文宽度，并指定文件名
    \vspace{-5pt}
    \caption{Ablation result on the AgiBot benchmark with and without in-domain EgoLive.} % 这里写图片的标题
    \label{fig:indomain_ablation} % 这里写引用标签，方便在正文中用 \ref{fig:indomain_ablation} 引用
    \vspace{-10pt}
\end{figure}

\textbf{RoboTwin 2.0 Simulation Benchmark.}
As shown in Table~\ref{tab:sub_human_data_ablation}, pretraining with the full EgoLive and JDAgibot datasets achieves the strongest overall result on RoboTwin 2.0, reaching an average success rate of $87.4\%$, outperforming both training from scratch and robot-only pretraining. This result indicates that large-scale egocentric human videos provide useful visual and behavioral priors for manipulation beyond what can be learned from robot trajectories alone. Compared to the full EgoLive setting, using only a 10\% subset does not yet provide the same benefit, suggesting that the contribution of EgoLive becomes more apparent when the dataset is used at sufficient scale. Notably, the improvement is not limited to a few isolated tasks, but is observed across a broad range of manipulation behaviors.

At the task level, as shown in Figure~\ref{fig:robotwin50_representative}, the benefit of EgoLive is most visible on tasks that require stronger spatial reasoning and multi-step structure, such as \textit{Stack Blocks Three}, \textit{Stack Bowls Three}, and \textit{Blocks Ranking RGB}. We also observe clear gains on complex interaction tasks such as \textit{Open Microwave}, as well as on fine-grained placement tasks such as \textit{Place Mouse Pad}, \textit{Place Cans Plasticbox}, and \textit{Place Fan}. These results suggest that EgoLive contributes transferable priors for object arrangement, sequential interaction, and precise placement, rather than helping only on a narrow subset of benchmark tasks. The comparison between the 10\% subset and the full EgoLive setting further reveals a clear positive scaling trend, with an approximately $6\%$ improvement in average success rate. This observation indicates that the benefit of human egocentric video does not saturate early, and supports our broader motivation that scaling diverse human-centric data is an effective way to improve transferable manipulation policies.

\textbf{Real-World AgiBot Benchmark.}
We further evaluate the effectiveness of EgoLive on the AgiBot G1 platform. As shown in Figure~\ref{fig:real_robot_evaluation_results}, incorporating human egocentric data consistently improves the real-robot performance of JoyAI-RA, increasing the average score from 0.56 to \textbf{0.74}. This result indicates that the benefit of EgoLive is not limited to simulation, but also transfers to downstream real-robot execution under practical deployment.

At the task level, human egocentric data yields the largest gains where semantic grounding and accurate object-to-target interaction matter most, notably on \textit{Headphones} and \textit{Remedy}, with smaller but consistent improvements on structured tasks such as \textit{Cup} and \textit{Croissant}, aligning with stronger instruction following, target identification, and multi-stage execution. The gains are not uniform: \textit{Mouse} and \textit{Food Scraps} remain comparatively weaker, suggesting that low-level execution sensitivity and coordination-heavy manipulation still depend on factors beyond scaling egocentric data alone. Overall, the ablation supports human egocentric data as an important contributor to JoyAI-RA's real-robot capability and to the system-level improvements.

\subsection{Can In-Domain EgoLive Data Enhance Model Generalization for Downstream Tasks?}

To investigate whether in-domain Egolive data benefits model generalization, we compare
two variants differing solely in the inclusion of in-domain demonstrations: (1)~\textbf{Baseline},
trained on the full data mixture including in-domain human videos; and (2)~\textbf{w/o In-Domain},
which excludes this subset while keeping all other data identical. Both variants use VLA pretraining only.
In-domain human video data is a subset of the broader human video corpus EgoLive,
consisting of demonstrations whose environments and task semantics broadly align with
the target robot deployment scenarios.

\input{table/ablation_nw-ego}

\begin{figure}[t]
    \centering
    \includegraphics[width=1.0\linewidth]{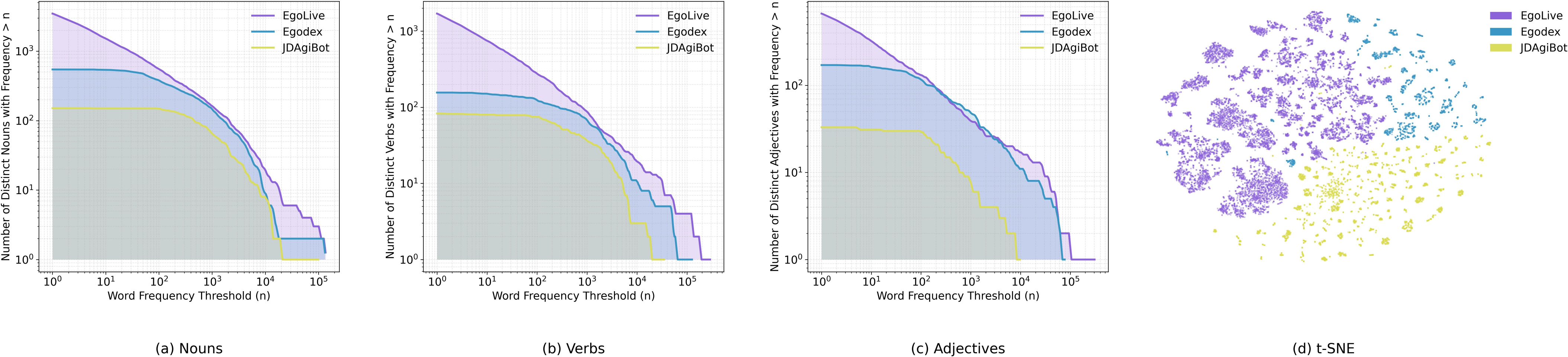}
    \vspace{-15pt}
    \caption{Data distribution comparison between Egolive and EgoDex~\cite{hoque2025egodex}. (a-c): Language instruction statistics across datasets. (d) t-SNE visualization of image features.}
    \label{fig:data_analysis}
    \vspace{-10pt}
\end{figure}

As shown in Figure~\ref{fig:indomain_ablation}, the \textbf{Baseline} achieves a higher
average success rate across the six tasks compared to \textbf{w/o In-Domain}, with a
modest overall gain. The Baseline outperforms on 4 of 6 tasks, with the most pronounced
improvements observed on \textit{Remedy} and \textit{Headphones}. We attribute these gains to a
strong alignment between the in-domain videos and the corresponding downstream tasks:
the collected demonstrations share similar object categories, spatial layouts, and
manipulation semantics with the evaluation scenarios, providing the model with
directly transferable visual and behavioral priors.

Despite the overall trend, the benefit is not consistent across all tasks. On \textit{Mouse} and \textit{Food Scraps}, \textbf{w/o
In-Domain} outperforms the Baseline. We attribute this to
distributional mismatch: the in-domain EgoLive for these tasks are collected
in settings that diverge from the evaluation scenarios in both environmental context
and task structure, introducing conflicting supervision signals rather than useful priors.

\subsection{Comparison of EgoLive and EgoDex Data}
\label{sec:ablation_data_source}

We compare two egocentric human video datasets, EgoLive and EgoDex~\cite{hoque2025egodex}, to analyze their effectiveness for VLA pretraining. As shown in Table~\ref{tab:ablation_data}, both datasets significantly improve performance over the baseline.
%when combined with Fourier-based retargeting.

When used individually, EgoDex achieves a success rate of 86.88, while EgoLive yields a slightly higher performance of 87.16. This suggests that both datasets provide useful behavioral priors for downstream manipulation, despite differences in scale and data distribution. Notably, EgoLive consists of long-horizon, continuous human interactions collected in real-world environments, which may provide richer temporal structure and more realistic motion patterns. In contrast, EgoDex contains a large number of segmented interaction clips with diverse behaviors, but its tasks are predominantly skewed toward toy-like manipulation.

To better understand this difference, we analyze the semantic properties of the datasets in Figure~\ref{fig:data_analysis}. As shown in Figure~\ref{fig:data_analysis}(a)--(c), our dataset exhibits a consistently heavier-tailed distribution across nouns, verbs, and adjectives, indicating broader vocabulary coverage and richer semantic diversity. In particular, the number of distinct words under higher frequency thresholds remains significantly larger, suggesting improved coverage of both common and long-tail concepts. Such diversity exposes the model to a wider range of objects, actions, and attributes, which is beneficial for generalization.

Furthermore, the T-SNE visualization in Figure~\ref{fig:data_analysis}(d) shows that our dataset spans a substantially larger and more continuous region in the semantic space, while EgoDex and JDAgiBot form more localized clusters. This indicates that EgoLive provides more diverse and less biased coverage of interaction semantics, which may contribute to improved robustness across tasks and environments.

When combining both datasets, performance further improves to 89.30, suggesting that they provide complementary information. Overall, these results highlight that both temporal structure and semantic diversity are critical factors in learning generalizable manipulation policies from human videos.

\subsection{Validity analysis of co-pretraining framework with human and simulation data}

\input{table/stage_and_sim_abl}

In this section, we conduct a comprehensive ablation study to validate the effectiveness of our two-stage co-pretraining framework (Stage 1: VLM Co-Pretraining; Stage 2: VLA Co-Pretraining) and the role of simulation data within the overall JoyAI-RA data proportion. As shown in Table ~\ref{tab:joyai_results}, the baseline model (VLA Post-Training only) achieves a success rate of 81.28\% on RoboTwin 2.0 (Easy). When we introduce only VLM co-pretraining or only VLA co-pretraining with human video data, the success rate improves to 87.84\% and 87.42\% respectively, demonstrating that each co-pretraining stage independently delivers meaningful performance gains. Combining both stages of co-pretraining with human data further elevates performance to 90.48\%, a 9.2 percentage point improvement over the baseline. 

In addition, we ablate the impact of simulation data in Stage 2: when simulation data is completely removed during VLA co-pretraining, performance drops by 1.14 percentage points (from 90.24\% to 89.10\%) relative to the full setup that includes simulation data. This performance gain stems from two key factors: first, simulation data naturally aligns with the visual distribution of downstream robotic manipulation tasks, reducing the distribution shift between pre-training and evaluation; second, it provides richer cross-embodiment action diversity during pre-training, enabling the model to generalize more effectively to unseen robot embodiments and manipulation scenarios. Together, these results confirm that both human video data and simulation data play complementary, critical roles in our co-pretraining framework, and highlight the importance of diverse in-domain data for robust robotic vision-language action learning.

%% file: table/sub_robotwin_full.tex
\definecolor{jdred}{RGB}{204,0,0}
\colorlet{jdredlight}{jdred!20}

\begin{table*}[t]
  \centering
  \footnotesize
  \setlength{\tabcolsep}{5.5pt}
  \caption{\textbf{Evaluation results on RoboTwin 2.0 benchmark.} The per-task results are shown in Appendix Table~\ref{tab:robotwin-short}.}
  \resizebox{\textwidth}{!}{%
  \begin{tabular}{
    *{1}{>{\raggedright\arraybackslash}m{3.5cm}} 
    *{8}{>{\centering\arraybackslash}m{0.75cm}} 
    *{2}{>{\centering\arraybackslash}m{0.95cm}}
  }
    \toprule
    \textbf{\multirow{2}{*}{Method}} 
      & \multicolumn{2}{c}{$\mathbf{\pi}_{0}$~\cite{black2410pi0}}
      & \multicolumn{2}{c}{$\mathbf{\pi}_{0.5}$~\cite{pi2025pi05}}
      & \multicolumn{2}{c}{Motus~\cite{bi2025motus}}
      & \multicolumn{2}{c}{\scriptsize{LingBot-VLA}~\cite{wu2026pragmatic}}
      & \multicolumn{2}{c}{\textbf{JoyAI-RA}} \\
    & Easy & Hard
    & Easy & Hard
    & Easy & Hard
    & Easy & Hard
    & Easy & Hard 
    \\
    \midrule
    \textbf{\textit{Success Rate(\%)}} &65.92&58.40&82.74&76.76&88.66&87.02&88.56&86.68&\textbf{90.48}&\textbf{89.28}\\
    \bottomrule
  \end{tabular}
  }
  \label{tab:sub-robotwin-short}
\vspace{-0.3cm}
\end{table*}

%% file: table/sub_robocasa.tex
\begin{table}[t]
\vspace{0.2cm}
\centering
\caption{\textbf{Evaluation results on RoboCasa GR1 Tabletop tasks.} The per-task results are shown in Appendix
Table~\ref{tab:robocasa-results}.}
\label{tab:sub-robocasa-results}
\setlength{\tabcolsep}{1pt}
\begin{adjustbox}{width=\textwidth}
\begin{tabular}{l>{\centering\arraybackslash}p{2.2cm}>{\centering\arraybackslash}p{1.6cm}>{\centering\arraybackslash}p{2.2cm}>{\centering\arraybackslash}p{2.3cm}>{\centering\arraybackslash}p{1.7cm}>{\centering\arraybackslash}p{1.7cm}>{\centering\arraybackslash}p{1.9cm}}
\toprule
\textbf{Method} &\scriptsize{GR00T-N1.6\cite{nvidia2025gr00t}} & \scriptsize{Qwen3PI\cite{community2026starvlalegolikecodebasevisionlanguageaction}} & \scriptsize{TwinBrainVLA}\scriptsize{\cite{yu2026twinbrainvlaunleashingpotentialgeneralist}} & \scriptsize{DualCoT-VLA}\scriptsize{\cite{zhong2026dualcotvlavisuallinguisticchainthought}} & \scriptsize{ABot-M0\cite{yang2026abot}} & \scriptsize{Being-H0.7\cite{beingh07}} & \scriptsize{\textbf{JoyAI-RA}}\\
\midrule
\textbf{Success Rate(\%)} & 47.6 & 43.9 & 54.6 & 55.1 &58.3 & 49.2 & \textbf{63.2}\\
\bottomrule
\end{tabular}
\end{adjustbox}
\vspace{-10pt}
\end{table}

%% file: table/sub_tasks_table.tex
\begin{table}[t]
  \centering
  \caption{\textbf{Egocentric human data ablation on RoboTwin 2.0 benchmark.} The full per-task results are provided in Appendix Table~\ref{tab:human_data_ablation}.}
  \vspace{-5pt}
  \begin{tabularx}{0.9\linewidth}{c *{4}{>{\centering\arraybackslash}X}}
    \toprule
    \textbf{Setting} 
    & \textbf{No Pretraining} 
    & \textbf{JDAgibot Only} 
    & \makecell{\textbf{EgoLive(10\%)}\\\textbf{$+$JDAgiBot}} 
    & \makecell{\textbf{EgoLive(Full)}\\\textbf{$+$JDAgiBot}} \\ 
    \midrule
    \textbf{\textit{Success Rate(\%)}} & 81.64 & 77.62 & 81.40 & \textbf{87.42} \\
    \bottomrule
  \end{tabularx}
  \label{tab:sub_human_data_ablation}
  \vspace{-5pt}
\end{table}

%% file: table/ablation_nw-ego.tex
\begin{table}[t]
\centering
\caption{Ablation on data sources on RoboTwin 2.0 (Easy) benchmark.}
\vspace{-5pt}
\label{tab:ablation_data}
\setlength{\tabcolsep}{5pt}
\renewcommand{\arraystretch}{1.15}

\begin{tabular}{lccccr}
\toprule
Setting & JDAgiBot & EgoDex & EgoLive & Success Rate $\uparrow$ \\
\midrule

Baseline & -- & -- & -- & 81.28 \\
+ JDAgiBot & $\checkmark$ & -- & -- & 79.20 \\
+ Human & $\checkmark$ & $\checkmark$ & $\checkmark$ & \textbf{89.30} \\

\midrule
\multicolumn{5}{l}{\textit{Single-source human data}} \\

Egodex only & -- & $\checkmark$ & -- & 86.88 \\
EgoLive only & -- & -- & $\checkmark$ & 87.16 \\

\bottomrule
\end{tabular}
\end{table}

%% file: table/stage_and_sim_abl.tex
\begin{table}[t]
\centering
\caption{Ablation experiments of RoboTwin 2.0 (Easy) benchmark on human data and simulation data across different training stages. Stage1: VLM Co-Pretraining; Stage2: VLA Co-Pretraining.}
\label{tab:joyai_results}
\begin{tabular}{lcc}
\toprule
\textbf{Training Paradigm} & \textbf{Training Configuration} & \textbf{Success Rate $\uparrow$}  \\
\midrule
Baseline(VLA Post-Training) & / & 81.28 \\
\multirow{1}{*}{Only VLM Co-Pretraining} 
& w human data & 87.84  \\
\multirow{1}{*}{Only VLA Co-Pretraining}
& w human data & 87.42  \\
\multirow{1}{*}{VLM + VLA Co-Pretraining}
& w human data (Stage 1 \& 2) & \textbf{90.48}  \\
\midrule
\multirow{2}{*}{VLM + VLA Co-Pretraining}
& w/o simulation data (Stage 2) & 89.10  \\
& w/ simulation data (Stage 2) & \textbf{90.24}  \\
\bottomrule
\end{tabular}
\vspace{-10pt} 
\end{table}

%% file: sections/conclusion.tex
\section{Conclusions}
\label{sec:conclusions}

We propose JoyAI-RA, a vision-language-action embodied foundation model for generalizable robotic manipulation. JoyAI-RA is built on a multi-source multi-level pretraining framework that jointly leverages web data, egocentric human manipulation videos, simulation-generated trajectories, and real-robot demonstrations. By structuring heterogeneous data according to embodiment proximity and scale, and by introducing action-space unification to bridge the embodiment gap, our approach enables effective transfer of manipulation knowledge across diverse data sources. Through comprehensive simulation and real-world testing, we validate the superiority of JoyAI-RA and conduct comprehensive ablation studies to analyze the effectiveness of the EgoLive data. Consequently, this suggests that a structured, multi-source approach to pretraining with action-space unification offers a viable trajectory toward truly generalizable embodied intelligence.

%% file: sections/appendix.tex
\appendix

\section*{Appendix}

\input{sections/contribution}

\renewcommand{\thefigure}{\Alph{figure}}
\renewcommand{\thetable}{\Alph{table}}

\setcounter{figure}{0}
\setcounter{table}{0}

\section{Real-World Experiment Details}
\label{sec:appendix_real_world}
\begin{itemize}
    \item \textbf{Task 1:} [Office] \textit{(Headphones) Headphones Hanging.} According to the instructions, the robot uses its right arm to pick up the headphones from the computer desk and hang them on the bracket. \\
    % \textit{Grading Rubric (\(K_t = 1\)):}
    \textit{Grading Rubric:}
        \begin{itemize}
            \item Pick up the headphones and hang them on the bracket.
        \end{itemize}
    \item \textbf{Task 2:} [Office] \textit{(Mouse) Mouse Picking and Placing.} According to the instructions, the robot uses its right arm to pick up the mouse from the computer desk and place it on the mouse pad. \\
    % \textit{Grading Rubric (\(K_t = 1\)):}
    \textit{Grading Rubric:}
        \begin{itemize}
            \item Pick up the mouse and place it on the mouse pad.
        \end{itemize}
    \item \textbf{Task 3:} [Tea Table] \textit{(Cup) Cup Discarding.} According to the instructions, the robot uses its right arm to pick up a paper cup from the table, pour the liquid into a container, and then throw the cup into the trash can. \\
    % \textit{Grading Rubric (\(K_t = 1\)):}
    \textit{Grading Rubric:}
        \begin{itemize}
            \item Pick up the paper cup, pour out the water, and put it in the trash can.
        \end{itemize}
    \item \textbf{Task 4:} [Kitchen] \textit{(Croissant) Croissant Toasting.} According to the instructions, the robot uses its right arm to open the oven, pulls out a baking tray, picks up a croissant from the plate, places it on the baking tray, pushes the baking tray into the oven, and then closes the oven. \\
    % \textit{Grading Rubric (\(K_t = 5\)):}
    \textit{Grading Rubric:}
        \begin{itemize}
            \item Open the oven door.
            \item Pull out the baking tray.
            \item Pick up the croissant and place it on the baking tray.
            \item Push the baking tray back in.
            \item Close the oven door.
        \end{itemize}
    \item \textbf{Task 5:} [Dining Table] \textit{(Food Scraps) Food Scraps Cleaning.} According to the instructions, the robot uses its right arm to pick up a plate containing food scraps, pours the scraps into a trash can, passes the plate to its left arm, and then places the plate on top of another plate. \\
    \textit{Grading Rubric:}
        \begin{itemize}
            \item Pick up the plate to empty food scraps.
            \item Pass the plate.
            \item Place the plate on top of the plate.
        \end{itemize}
    \item \textbf{Task 6:} [Pharmacy] \textit{(Remedy) Remedy Packaging.} According to the instructions, the robot uses its left arm to place the package in the center of the table, then uses its right arm to pick out the designated drug from among many medicines and put it into the package. \\
    \textit{Grading Rubric:}
        \begin{itemize}
            \item Pick up and place the package bag.
            \item Pick up and put the remedy into the package bag.
        \end{itemize}
\end{itemize}

\section{Experiment Results}
\label{sec:appendix_exp}

The full table of Table~\ref{tab:sub-robotwin-short}, Table~\ref{tab:sub-robocasa-results} and Table~\ref{tab:sub_human_data_ablation} are shown in the Table~\ref{tab:robotwin-short}, Table~\ref{tab:robocasa-results} and Table~\ref{tab:human_data_ablation}.

\input{table/robotwin_full}

\input{table/robocasa}

\input{table/tasks_table}

%% file: sections/contribution.tex
\section{Contributions}
\label{sec:contributions}

\begin{itemize}[left=0pt]
    \item \textbf{Core Contributors:} \\
    Tianle Zhang, Zhihao Yuan, Dafeng Chi, Peidong Liu, Dongwei Li, Kejun Hu, Likui Zhang, Junnan Nie, Ziming Wei, Zengjue Chen, Yili Tang, Jiayi Li, Zhiyuan Xiang, Mingyang Li, Tianci Luo, Hanwen Wan, Ao Li, Linbo Zhai, Zhihao Zhan, Yuzheng Zhuang\textsuperscript{\dagger}, Liang Lin\textsuperscript{\dagger}
    \item \textbf{Contributors (alphabetical order by last names):} \\
    Xiaodong Bai, Jiakun Cai, Peng Cao, Kangliang Chen, Siang Chen, Yixiang Dai, Shuai Di, Nan Duan, Yicheng Gong, Chenguang Gui, Yucheng Guo, Peng Hao, Qingrong He, Haoyang Huang, Kunrui Huang, Zhixuan Huang, Shibo Jin, Yixiang Jin, Anson Li, Dongjiang Li, Jiawei Li, Ruodai Li, Yihang Li, Yuzhen Li, Jiaming Liang, Fangsheng Liu, Jing Long, Mingxi Luo, Xing Pan, Hui Shen, Xiaomeng Tian, Daming Wang, Song Wang, Junwu Xiong, Hang Xu, Wanting Xu, Zhengcheng Yu, He Zhang, Jiyao Zhang, Lin Zhao, Chen Zhou

\end{itemize}

\renewcommand{\thefootnote}{\dagger}
\footnotetext{Corresponding author:Yuzheng Zhuang<zhuangyuzheng.1@jd.com>, Liang Lin<linliang@ieee.org>.}

%% file: table/robotwin_full.tex
\definecolor{jdred}{RGB}{204,0,0}
\colorlet{jdredlight}{jdred!20}

\begin{table*}[htbp]
  \centering
  \footnotesize
  \setlength{\tabcolsep}{5.5pt}
  \caption{\textbf{Full evaluation results on RoboTwin 2.0 benchmark.}}
  \begin{tabular}{
    *{1}{>{\centering\arraybackslash}m{3.5cm}} 
    *{8}{>{\centering\arraybackslash}m{0.75cm}} 
    *{2}{>{\columncolor{jdredlight}\centering\arraybackslash}m{0.95cm}}
  }
    \toprule
    \textbf{\multirow{2}{*}{Simulation Task}} 
      & \multicolumn{2}{c}{$\mathbf{\pi}_{\mathbf{0}}$~\cite{black2410pi0}}
      & \multicolumn{2}{c}{$\mathbf{\pi}_{\mathbf{0.5}}$~\cite{pi2025pi05}}
      & \multicolumn{2}{c}{Motus~\cite{bi2025motus}}
      & \multicolumn{2}{c}{\scriptsize{LingBot-VLA}~\cite{wu2026pragmatic}}
      & \multicolumn{2}{c}{\cellcolor{jdredlight}{\textbf{JoyAI-RA}}} \\
    & \textbf{Easy} & \textbf{Hard} 
    & \textbf{Easy} & \textbf{Hard} 
    & \textbf{Easy} & \textbf{Hard} 
    & \textbf{Easy} & \textbf{Hard} 
    & \textbf{Easy} & \textbf{Hard} 
    \\
    \midrule
\textit{Adjust Bottle}&99&95&100&99&89&93&100&100&100&100\\
\textit{Beat Block Hammer}&79&84&96&93&95&88&92&89&95&91\\
\textit{Blocks Ranking RGB}&80&63&92&85&99&97&92&91&94&93\\
\textit{Blocks Ranking Size}&14&5&49&26&75&63&76&70&81&75\\
\textit{Click Alarmclock}&77&68&98&89&100&100&97&43&64&56\\
\textit{Click Bell}&71&48&99&66&100&100&43&36&81&70\\
\textit{Dump Bin Bigbin}&88&83&92&97&95&91&97&97&97&99\\
\textit{Grab Roller}&98&94&100&100&100&100&100&100&100&100\\
\textit{Handover Block}&47&31&66&57&86&73&99&93&99&93\\
\textit{Handover Mic}&97&97&98&97&78&63&100&99&100&99\\
\textit{Hanging Mug}&14&11&18&17&38&38&31&28&31&28\\
\textit{Lift Pot}&80&72&96&85&96&99&100&99&100&99\\
\textit{Move Can Pot}&68&48&51&55&34&74&97&87&97&87\\
\textit{Move Pillbottle Pad}&67&46&84&61&93&96&98&99&98&99\\
\textit{Move Playingcard Away}&74&65&96&84&100&96&99&95&99&95\\
\textit{Move Stapler Pad}&41&24&56&42&83&85&93&96&93&96\\
\textit{Open Laptop}&71&81&90&96&95&91&96&100&96&100\\
\textit{Open Microwave}&4&32&34&77&95&91&97&99&97&99\\
\textit{Pick Diverse Bottles}&69&31&81&71&90&91&85&90&85&90\\
\textit{Pick Dual Bottles}&59&37&93&63&96&90&95&93&95&93\\
\textit{Place A2B Left}&43&47&87&82&82&79&99&96&99&96\\
\textit{Place A2B Right}&39&34&87&84&90&87&97&92&97&92\\
\textit{Place Bread Basket}&62&46&77&64&91&94&88&91&88&91\\
\textit{Place Bread Skillet}&66&49&85&66&86&83&92&89&92&89\\
\textit{Place Burger Fries}&81&76&94&87&98&98&99&93&99&93\\
\textit{Place Can Basket}&55&46&62&62&81&76&71&73&71&73\\
\textit{Place Cans Plasticbox}&63&45&94&84&98&94&100&98&100&98\\
\textit{Place Container Plate}&97&92&99&95&98&99&96&99&96&99\\
\textit{Place Dual Shoes}&59&51&75&75&93&87&90&97&90&97\\
\textit{Place Empty Cup}&91&85&100&99&99&98&100&100&100&100\\
\textit{Place Fan}&66&71&87&85&91&87&91&92&91&92\\
\textit{Place Mouse Pad}&20&20&60&39&66&68&89&82&89&82\\
\textit{Place Object Basket}&67&70&80&76&81&87&90&88&90&88\\
\textit{Place Object Scale}&57&52&86&80&88&85&90&87&90&87\\
\textit{Place Object Stand}&82&68&91&85&98&97&95&93&95&93\\
\textit{Place Phone Stand}&49&53&81&81&87&86&95&95&95&95\\
\textit{Place Shoe}&76&76&92&93&99&97&99&100&99&100\\
\textit{Press Stapler}&44&37&87&83&93&98&87&81&87&81\\
\textit{Put Bottles Dustbin}&65&56&84&79&81&79&95&97&95&97\\
\textit{Put Object Cabinet}&73&60&80&79&88&71&87&86&87&86\\
\textit{Rotate QRcode}&74&70&89&87&89&73&83&82&83&82\\
\textit{Scan Object}&55&42&72&65&67&66&98&96&98&96\\
\textit{Shake Bottle Horizontally}&98&92&99&99&100&98&100&100&100&100\\
\textit{Shake Bottle}&94&91&99&97&100&97&100&100&100&100\\
\textit{Stack Blocks Three}&72&52&91&76&91&95&60&62&60&62\\
\textit{Stack Blocks Two}&93&79&97&100&100&98&95&93&95&93\\
\textit{Stack Bowls Three}&77&75&77&71&79&87&80&81&80&81\\
\textit{Stack Bowls Two}&94&95&95&96&98&98&95&93&95&93\\
\textit{Stamp Seal}&46&33&79&55&93&92&90&90&90&90\\
\textit{Turn Switch}&41&42&62&54&84&78&71&76&71&76\\
    \midrule
    \textbf{\textit{Average(\%)}} &65.92&58.40&82.74&76.76&88.66&87.02&88.56&86.68&\textbf{90.48}&\textbf{89.28}\\
    \bottomrule
  \end{tabular}
  \label{tab:robotwin-short}
\vspace{-0.3cm}
\end{table*}

%% file: table/robocasa.tex
\begin{table}[htbp]
\centering
\caption{\textbf{Evaluation results on RoboCasa GR1 Tabletop tasks.}}
\label{tab:robocasa-results}
\setlength{\tabcolsep}{1pt}
\begin{adjustbox}{width=\textwidth}
\begin{tabular}{l>{\centering\arraybackslash}p{2.5cm}>{\centering\arraybackslash}p{2.25cm}>{\centering\arraybackslash}p{2.25cm}>{\centering\arraybackslash}p{2.7cm}>{\centering\arraybackslash}p{2.25cm}>{\columncolor{jdredlight}\centering\arraybackslash}p{2.25cm}}
\toprule
\textbf{Task} &\small{GR00T-N1.6\cite{nvidia2025gr00t}} & \small{Qwen3PI\cite{community2026starvlalegolikecodebasevisionlanguageaction}} & \scriptsize{TwinBrainVLA}\small{\cite{yu2026twinbrainvlaunleashingpotentialgeneralist}} & \scriptsize{DualCoT-VLA}\small{\cite{zhong2026dualcotvlavisuallinguisticchainthought}} & \small{ABot-M0\cite{yang2026abot}} & \small{\textbf{JoyAI-RA}}\\
\midrule
BottleToCabinetClose & 51.5 & 26.0 & 74.0 & 66.0 &86.0 & 84.0\\
CanToDrawerClose & 13.0 & 62.0 & 72.0 & 64.0 &74.0 & 90.0\\
CupToDrawerClose & 8.5 & 42.0 & 52.0 & 46.0 &48.0 & 48.0\\
MilkToMicrowaveClose & 14.0 & 50.0 & 60.0 & 58.0 &46.0 & 84.0\\
PotatoToMicrowaveClose & 41.5 & 42.0 & 36.0 & 30.0 &50.0 & 70.0\\
WineToCabinetClose & 16.5 & 32.0 & 46.0 & 38.0 &66.0 & 54.0\\
\addlinespace
CuttingboardToBasket & 58.0 & 40.0 & 62.0 & 44.0 &70.0 & 88.0\\
CuttingboardToCardboardbox & 46.5 & 46.0 & 46.0 & 54.0 &58.0 & 46.0\\
CuttingboardToPan & 68.5 & 60.0 & 70.0 & 80.0 &76.0 & 92.0\\
CuttingboardToPot & 65.0 & 40.0 & 66.0 & 64.0 &66.0 & 80.0\\
CuttingboardToTieredbasket & 46.5 & 44.0 & 52.0 & 46.0 &38.0 & 36.0\\
\addlinespace
PlacematToBasket & 58.5 & 44.0 &30.0 & 48.0 &52.0 & 76.0\\
PlacematToBowl & 57.5 & 52.0 & 54.0 & 58.0 &66.0 & 52.0\\
PlacematToPlate & 63.0 & 50.0 & 64.0 & 74.0 &60.0 & 38.0\\
PlacematToTieredshelf & 28.5 & 28.0 & 38.0 & 26.0 &26.0 & 14.0\\
\addlinespace
PlateToBowl & 57.0 & 52.0 & 60.0 & 50.0 &54.0 & 48.0\\
PlateToCardboardbox & 43.5 & 40.0 & 58.0 & 56.0 &48.0 & 38.0\\
PlateToPan & 51.0 & 36.0 & 56.0 & 70.0 &66.0 & 46.0\\
PlateToPlate & 78.7 & 48.0 & 66.0 & 76.0 &64.0 & 88.0\\
\addlinespace
TrayToCardboardbox & 51.5 & 34.0 & 46.0 & 52.0 &54.0 & 82.0\\
TrayToPlate & 71.0 & 64.0 & 72.0 & 64.0 &68.0 & 88.0\\
TrayToPot & 64.5 & 44.0 & 56.0 & 70.0 &64.0 & 88.0\\
TrayToTieredbasket & 57.0 & 50.0 & 46.0 & 60.0 &60.0 & 62.0\\
TrayToTieredshelf & 31.5 & 28.0 & 28.0 & 28.0 &38.0 & 24.0\\
\midrule
\textbf{Average(\%)} & 47.6 & 43.9 & 54.6 & 55.1 &58.3 & \textbf{63.2}\\
\bottomrule
\end{tabular}
\end{adjustbox}
\end{table}

%% file: table/tasks_table.tex
\begin{table}[htbp]
  \centering
  \caption{\textbf{Egocentric human data ablation on RoboTwin 2.0 benchmark.}}
  \small
  \begin{tabularx}{\linewidth}{l *{4}{>{\centering\arraybackslash}X}}
    \toprule
    \textbf{Simulation Task} 
    & \textbf{No Pretrain} 
    & \textbf{JDAgibot Only} 
    & \makecell{\textbf{EgoLive(10\%)}\\\textbf{$+$JDAgiBot}} 
    & \makecell{\textbf{EgoLive(Full)}\\\textbf{$+$JDAgiBot}} \\
    \midrule
    \textit{Adjust Bottle} & 100 & 100 & 100 & 100 \\
    \textit{Beat Block Hammer} & 96 & 90 & 85 & 86 \\
    \textit{Blocks Ranking RGB} & 74 & 68 & 61 & 99 \\
    \textit{Blocks Ranking Size} & 70 & 38 & 62 & 80 \\
    \textit{Click Alarmclock} & 2 & 3 & 7 & 11 \\
    \textit{Click Bell} & 40 & 50 & 41 & 59 \\
    \textit{Dump Bin Bigbin} & 96 & 89 & 98 & 98 \\
    \textit{Grab Roller} & 100 & 100 & 100 & 100 \\
    \textit{Handover Block} & 94 & 88 & 86 & 73 \\
    \textit{Handover Mic} & 85 & 94 & 94 & 96 \\
    \textit{Hanging Mug} & 39 & 27 & 31 & 38 \\
    \textit{Lift Pot} & 99 & 0 & 0 & 100 \\
    \textit{Move Can Pot} & 92 & 95 & 95 & 87 \\
    \textit{Move Pillbottle Pad} & 96 & 92 & 95 & 95 \\
    \textit{Move Playingcard Away} & 100 & 100 & 100 & 100 \\
    \textit{Move Stapler Pad} & 73 & 79 & 81 & 71 \\
    \textit{Open Laptop} & 94 & 98 & 99 & 95 \\
    \textit{Open Microwave} & 87 & 44 & 89 & 100 \\
    \textit{Pick Diverse Bottles} & 84 & 86 & 88 & 76 \\
    \textit{Pick Dual Bottles} & 95 & 94 & 99 & 91 \\
    \textit{Place A2B Left} & 86 & 98 & 94 & 93 \\
    \textit{Place A2B Right} & 65 & 93 & 95 & 94 \\
    \textit{Place Bread Basket} & 88 & 80 & 84 & 88 \\
    \textit{Place Bread Skillet} & 88 & 90 & 87 & 88 \\
    \textit{Place Burger Fries} & 98 & 91 & 98 & 98 \\
    \textit{Place Can Basket} & 66 & 70 & 79 & 73 \\
    \textit{Place Cans Plasticbox} & 88 & 82 & 96 & 99 \\
    \textit{Place Container Plate} & 98 & 100 & 100 & 99 \\
    \textit{Place Dual Shoes} & 82 & 84 & 82 & 89 \\
    \textit{Place Empty Cup} & 98 & 92 & 99 & 100 \\
    \textit{Place Fan} & 84 & 87 & 92 & 94 \\
    \textit{Place Mouse Pad} & 73 & 65 & 67 & 87 \\
    \textit{Place Object Basket} & 86 & 74 & 88 & 89 \\
    \textit{Place Object Scale} & 85 & 91 & 94 & 89 \\
    \textit{Place Object Stand} & 96 & 96 & 95 & 92 \\
    \textit{Place Phone Stand} & 76 & 91 & 81 & 91 \\
    \textit{Place Shoe} & 100 & 97 & 98 & 97 \\
    \textit{Press Stapler} & 95 & 81 & 94 & 91 \\
    \textit{Put Bottles Dustbin} & 90 & 91 & 89 & 90 \\
    \textit{Put Object Cabinet} & 45 & 83 & 80 & 93 \\
    \textit{Rotate QRcode} & 86 & 76 & 86 & 89 \\
    \textit{Scan Object} & 93 & 87 & 91 & 89 \\
    \textit{Shake Bottle Horizontally} & 100 & 100 & 100 & 100 \\
    \textit{Shake Bottle} & 99 & 100 & 100 & 99 \\
    \textit{Stack Blocks Three} & 40 & 28 & 34 & 89 \\
    \textit{Stack Blocks Two} & 84 & 81 & 80 & 99 \\
    \textit{Stack Bowls Three} & 61 & 29 & 43 & 89 \\
    \textit{Stack Bowls Two} & 79 & 70 & 80 & 94 \\
    \textit{Stamp Seal} & 71 & 72 & 87 & 73 \\
    \textit{Turn Switch} & 66 & 67 & 66 & 61 \\
    \midrule
    \textbf{\textit{Average(\%)}} & 81.64 & 77.62 & 81.40 & 87.42 \\
    \bottomrule
  \end{tabularx}
  \label{tab:human_data_ablation}
\end{table}